# Impact of Tuning Parameters in Deep Convolutional Neural Network Using a Crack Image Dataset

**Mahe Zabin[1], Ho-Jin Choi[1*], Md. Monirul Islam[2], and Jia Uddin[3]**

[1]School of Computing, Korea Advanced Institute of Science & Technology, KAIST, Daejeon, 34141, South Korea
[2]Department of Textile Engineering, Uttara University, Dhaka, 1230, Bangladesh
[3]AI and Big Data Department, Woosong University, Daejeon, 300718, South Korea
{mahezabin@kaist.ac.kr, hojinc@kaist.ac.kr, monirul@uttarauniversity.edu.bd, jia.uddin@wsu.ac.kr}

**Abstract.** The performance of a classifier depends on the tuning of its parameters. In this paper, we have experimented the impact of various tuning parameters on the performance of a deep convolutional neural network (DCNN). In the experimental evaluation, we have considered a DCNN classifier that consists of 2 convolutional layers (CL), 2 pooling layers (PL), 1 dropout, and a dense layer. To observe the impact of pooling, activation function, and optimizer tuning parameters, we utilized a crack image dataset having two classes: negative and positive. The experimental results demonstrate that with the maxpooling, the DCNN demonstrates its better performance for adam optimizer and tanh activation function.

## 1 Introduction

A Deep CNN (Deep Convolutional Neural Network) usually consists of several neurons layers, each layer being a non-linear operation on the linear transformation of the output of the previous layer [1]. CLs (Convolution Layers) and PLs (Pooling Layers) are the most common layers in a Deep CNN. The CLs have weights that must be learned, but the PLs use a fixed function to convert the activation. The performance of a CNN model depends on the activation functions, optimizers algorithms, and various PLs. The activation function is both a decision-making function and support for learning complex patterns. Choosing the right activation function can speed up the learning process.

In [2-3], the authors discussed the performances of models using activation functions. In [4], the authors have utilized 3 optimizers- sgdm, adam, rmsprop for the performance of the deep CNN. In [5], the authors utilized adam activation and softmax optimizers for the classification of the traffic sign. The authors described the tanh function for the hardware friendly in the CNNs [6]. In [7], a hybrid pooling layer including max and average pooling is used for the Cherry fruit classification using the deep CNN model. In [8], an ECG signal classification model is proposed using deep CNN and adadelta optimizer. In [9], the softplus activation function is used for the CIFAR-10 dataset. For super-resolution pictures, 4 optimization techniques (SGD, AdaGrad, RMSP, and ADAM) are used in [10]; where Adam optimizer performs well than the others. In [11], a CNN-based feature extractor using the Adagrad optimizer is used to extract the feature vectors from the input images.

\* Corresponding Author

In this paper, we have considered 5 gradient-based optimizers (Stochastic Gradient Descent (SGD), Adaptive Gradient (AdaGrad), Adaptive Delta (AdaDelta), Root Mean Square Propagation (RMSProp), and Adaptive Moment Estimate (Adam)), 8 activation functions (relu, tanh, gelu, elu, selu, silu, softmax, along with softplus), and 2 poolings (max and average) as tuning parameters.

The remainder of the paper is as follows. Chapter 2 includes the methodology. The experimental result analysis is in Chapter 3. Finally, conclude the paper is in Chapter 4.

## 2    Methodology

Figure 1 shows a detailed block diagram of the methodology. We used various activation functions in the 2-D CLs. As PLs, we applied maxpooling and average pooling. In addition, we have used 5 optimizers to check the performance of individual optimized for the crack image dataset.

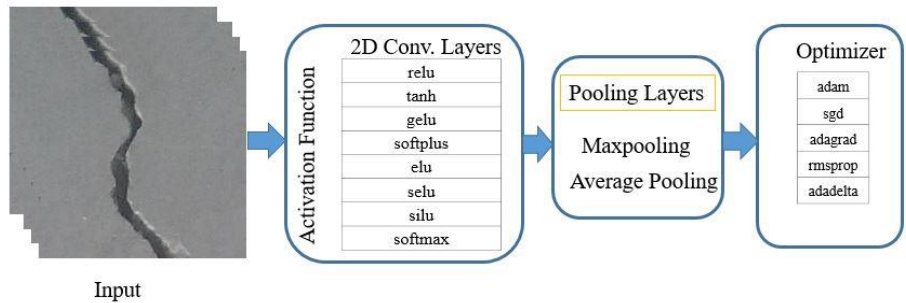

**Fig. 1.** A block diagram of the methodology

### 2.1 Dataset

To evaluate the performance of different tuning parameters we have used a crack image dataset [12] and a few sample images are shown in Fig. 2. The dataset consists of 2 classes- negative and positive crack images. In the evaluation, we have used a total of 800 images (400 negative, 400 positive) with a size 227x227 with RGB channels.

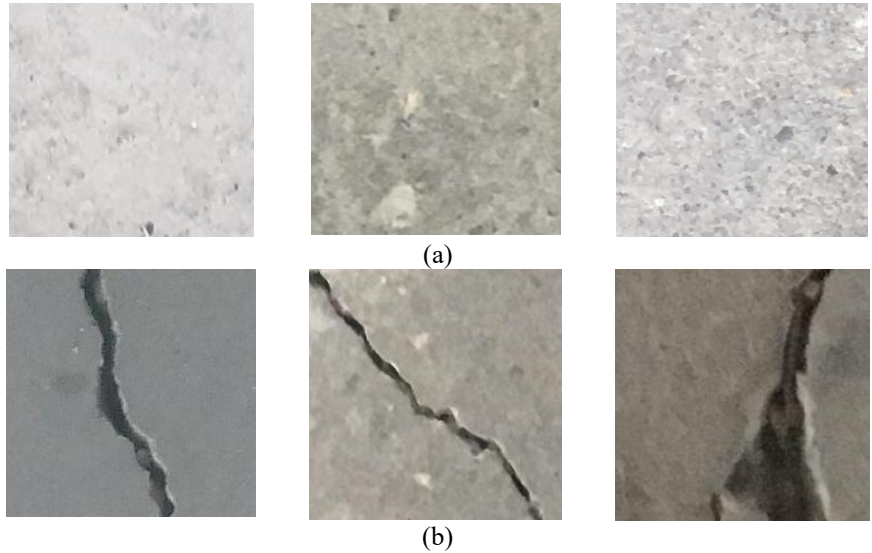

**Fig. 2.** Crack image dataset (a) negative, (b) positive.

## 2.2 Convolutional Layers

The first layer of CNN is the CL. It consists of a set of convolutional kernels where all neurons act as a kernel. The kernel works by dividing the image into small slices. To extract the feature, an image is split into a number of small blocks. Kernel multiplies its elements by corresponding field elements using a set of weights [13]. To increase the performance of CL, linear filter is used with the non-linear function [14].

An activation function is used to learn the pattern of data in the CL. The following snapshot of the code is related to the activation function.

```
model.add(Conv2D(256, (3, 3),input_shape=X.shape[1:]))
            model.add(Activation(X))
```

Where X stands for the activation function which can be either non-linear or linear.

## 2.3 Pooling Layers

In general, a pooling layer lies between two successive CLs. The PL decreases the number of parameters as well as computation by down-sampling the representation. There are various pooling functions in deep learining including max (1-3)D, average (1-3)D, globalmax (1-3)D, globalaverage (1-3)D [14, 16-17].

The following snapshot is related to max pooling, where the pooling size is 2x2.

```
model.add(MaxPooling2D(pool_size=(2, 2)))
```

## 2.4 Optimizers

In deep learning, optimizers are used to reduce the loss function and update the weights in the backpropagation [4]. Several optimizers are- Gradient Descent-SGD [21], Adaptive Gradient-AdaGrad, Adaptive Delta-AdaDelta, Root Mean Square Propagation-RMSProp, and Adaptive Moment Estimate-Adam. To remove the problem (very slow during small gradient, gradient-dependent updation at every iteration, and following the erroneous gradient frequently for the noisy gradients) of SGD, AdaGrad is proposed [18]. There are two problems in AdaGrad including decreasing learning rate, and a requirement to pick the global learning rate. AdaDelta is proposed for fixing these issues [19]. RMSProp is presented for almost the same answer the AdaGrad's first problem. Two techniques named RMSprop, and SGD are combinedly presented in Adam optimizer [20].

```
model.compile(loss='binary_crossentropy',
              optimizer='Y',
   metrics=['accuracy', 'MeanSquaredError', 'AUC'])
```

In the snapshot, Y stands for an optimizer.

## 3 Experimental Result Analysis

In this section, we present the analysis of different activation functions, pooling, and optimizers. As a performance metrics, we have used accuracy both in validation and training. Accuracy can be defined from the confusion matrix which is an N × N matrix used for assessing the classification model, where N is the number of goal classes [15]. An overall sight of the confusion matrix is shown in Table I.

TABLE I. CONFUSION MATRIX

|  | Predicted Yes | Predicted No |
|---|---|---|
| Actual Yes | Tp | fn |
| Actual No | Fp | tn |

where *tp* is the true positive number of positive events of a data set that are correctly predicted, tn refers to the number of negative events of a data set that are correctly predicted, *fp* is the measure of positive cases that are predicted incorrectly, and *fn* is the measure of negative cases that are predicted incorrectly. We can define the accuracy mathematically by using equation 1.

$$\text{Accuracy} = \frac{tp + tn}{tp + tn + fp + fn} \quad (1)$$

As depicted in Table II, the max-pooling works better in all cases. That is why we choose max-pooling in the next experimental steps. The table also shows the Adam optimizer demonestrated the highest accuracy both in training and validation.

TABLE II.  EXPERIMENTAL SUMMARY

| Activation | Optimizer | Max Pooling Accuracy | | Average Pooling Accuracy | |
|---|---|---|---|---|---|
| | | Training | Validation | Training | Validation |
| Relu | adam | 0.9181 | 0.9667 | 0.9875 | 0.9458 |
| tanh | | **1** | **0.9542** | 1 | 0.9083 |
| gelu | | 0.9964 | 0.9667 | 0.9571 | 0.9042 |
| softplus | | 0.4661 | 0.4958 | 0.5071 | 0.4833 |
| elu | | 0.9982 | 0.9625 | 0.9679 | 0.8917 |
| selu | | 0.9929 | 0.95 | 0.9768 | 0.8917 |
| silu | | 0.9929 | 0.9458 | 0.9571 | 0.9125 |
| softmax | | 0.8679 | 0.8667 | 0.8554 | 0.8458 |
| Relu | Sgd | 0.8643 | 0.8625 | 0.5071 | 0.4833 |
| tanh | | 0.9232 | 0.9 | 0.8982 | 0.8625 |
| gelu | | 0.8554 | 0.8792 | 0.8643 | 0.8708 |
| softplus | | 0.4554 | 0.4958 | 0.5625 | 0.5167 |
| elu | | 0.925 | 0.9167 | 0.8982 | 0.8583 |
| selu | | 0.9464 | 0.9292 | 0.9143 | 0.8625 |
| silu | | 0.8589 | 0.8667 | 0.8571 | 0.8708 |
| softmax | | 0.8786 | 0.875 | 0.8661 | 0.8708 |
| Relu | adagrad | 0.9 | 0.8917 | 0.5071 | 0.4833 |
| tanh | | 0.8857 | 0.875 | 0.8804 | 0.8208 |
| gelu | | 0.7196 | 0.6583 | 0.5554 | 0.625 |
| softplus | | 0.5125 | 0.4958 | 0.6107 | 0.8625 |
| elu | | 0.8625 | 0.9042 | 0.5268 | 0.4833 |
| selu | | 0.8982 | 0.9 | 0.8804 | 0.8583 |
| silu | | 0.6429 | 0.6292 | 0.6393 | 0.5792 |
| softmax | | 0.5018 | 0.4958 | 0.5071 | 0.4833 |
| Relu | rmsprop | 0.9946 | 0.95 | 0.9661 | 0.9083 |
| tanh | | 0.963 | 0.95 | 0.7518 | 0.8458 |
| gelu | | 0.984 | 0.946 | 0.8839 | 0.8917 |
| softplus | | 0.55 | 0.496 | 0.7536 | 0.7583 |
| elu | | 0.941 | 0.958 | 0.8589 | 0.5167 |
| selu | | 0.97 | 0.942 | 0.9214 | 0.7708 |
| silu | | 0.986 | 0.963 | 0.8839 | 0.8958 |
| softmax | | 0.491 | 0.496 | 0.7661 | 0.7667 |
| Relu | adadelta | 0.9964 | 0.9583 | 0.4929 | 0.5167 |
| tanh | | 0.7607 | 0.7833 | 0.5036 | 0.5083 |
| gelu | | 0.4982 | 0.5042 | 0.4982 | 0.5042 |
| softplus | | 0.4946 | 0.5042 | 0.4893 | 0.4833 |
| elu | | 0.5304 | 0.5167 | 0.5429 | 0.5042 |
| selu | | 0.7179 | 0.6667 | 0.6321 | 0.5833 |
| silu | | 0.4982 | 0.5042 | 0.4982 | 0.5042 |
| softmax | | 0.5018 | 0.4958 | 0.5071 | 0.4833 |

It has been observed that the highest training accuracy and validation accuracy have been exhibited for tanh and relu activation function, respectively. The reason why tanh is better for our implementation is we are considering binary classes. For hyperbolic tangent (tanh), the range is between -1 and 1 compared to 0 and 1. This makes the function to be more convenient for neural networks. Relu performs better in tems of validation accuracy because tanh might lead to fading gradient problem. This means if

*x* is smaller than -2 or bigger than 2, the derivative gets really small and your network might not converge, or you might end up having a dead neuron that does not fire anymore.

### 3.1 Impact of pooling on the performance during activation and optimizer as constant

In Table III, we see that the max-pooling demonstrates the highest score when the activation function and optimizer are the same. Therefore, we have chosen, max-pooling for the next experiments. Average pooling sustains a lot of data. Average pooling sometimes cannot extract the crucial features as it considers everything and gives an average value which may or may not be important. On the other hand, max pooling rejects a good amount of data by extracting only the most salient features. Max pooling focusses only on the important features. Average pooling encourages the network to identify the complete extent of the object, whereas max pooling restricts that to only the very important features. This is one of the primary reasons that max pooling performs better however, choice in the pooing layer always depends on what we expect from the CNN.

TABLE III. IMPACT OF POOLING DURING TANH AND ADAM

| Pooling | Activation | Optimizer | Accuracy | |
|---|---|---|---|---|
| | | | Training | Validation |
| Max | tanh | Adam | **1** | **0.9542** |
| Average | tanh | Adam | 1 | 0.9083 |

### 3.2 Impact of activation function during max pooling and adam optimizer

In Table IV, we can see the tanh function exhibited the highest score for the max pooling and Adam optimizer. So, we have chosen, tanh function for the next experiments.

TABLE IV. IMPACT OF ACTIVATION DURING MAX AND ADAM

| Pooling | Activation | Optimizer | Accuracy | |
|---|---|---|---|---|
| | | | Training | Validation |
| Max | Relu | Adam | 0.9181 | 0.9667 |
| | Tanh | | **1** | **0.9542** |
| | Gelu | | 0.9964 | 0.9667 |
| | Softplus | | 0.4661 | 0.4958 |
| | Elu | | 0.9982 | 0.9625 |
| | Selu | | 0.9929 | 0.95 |
| | Silu | | 0.9929 | 0.9458 |
| | Softmax | | 0.8679 | 0.8667 |

### 3.3 Impact of optimizers during max pooling and activation functions

In depicted in Table V, the adam optimizer demonestrated the highest classification accuracy for the tanh activation function and max-pooling. It is mainly because Adam can handle sparse gradients on noisy problems. It is the combination of the best properties of the AdaGrad and RMSProp algorithms.

TABLE V. IMPACT OF OPTIMIZER DURING MAX AND TANH

| Pooling | Activation | Optimizer | Accuracy | |
|---|---|---|---|---|
| | | | Training | Validation |
| Max | tanh | Adam | **1** | **0.9542** |
| | | Sgd | 0.9232 | 0.90 |
| | | Adagrad | 0.8857 | 0.875 |
| | | Rmsprop | 0.963 | 0.95 |
| | | Adadelta | 0.7607 | 0.7833 |

# 4 Conclusion

This paper studied the impact of various tuning parameters on the performance of a deep convolutional neural network. For the 8 activation functions (relu, selu, silu, tanh, elu, softmax, softplus, and gelu), 5 optimizers (adam, adagrad, adadelta, sgd, and rmsprop), and 2 pooling (max and average), we have observed the highest training accuracy 1 and validation accuracy 0.9542 for the max pooling, adam optimizer, and tanh function for the crack image dataset having 2 classes. In the future, the impact of more tuning parameters like kernel size of convolutional and pooling layers, drop-out rate may also check for other benchmark datasets.

**Acknowledgment**
This research was supported and funded by the Korean National Police Agency. [Project Name: XR Counter-Terrorism Education and Training Test Bed Establishment/Project Number: PR08-04-000-21]